\documentclass{article}

\usepackage{microtype}
\usepackage{graphicx}
\usepackage{subfigure}
\usepackage{booktabs} 

\usepackage{hyperref}

\usepackage[utf8]{inputenc} 
\usepackage{amssymb}
\usepackage{float}
\usepackage{amsmath}
\usepackage{bm}
\usepackage{multirow}

\usepackage[accepted]{icml2019}

\icmltitlerunning{\small Robust Sampling in Deep Learning}

\begin{document}

\twocolumn[
\icmltitle{Robust Sampling in Deep Learning}



\icmlsetsymbol{equal}{*}

\begin{icmlauthorlist}
\icmlauthor{Aurora Cobo Aguilera}{uc3m}
\icmlauthor{Antonio Artés-Rodríguez}{uc3m}
\icmlauthor{Fernando Pérez-Cruz}{eth}
\icmlauthor{Pablo Martínez Olmos}{uc3m}
\end{icmlauthorlist}

\icmlaffiliation{uc3m}{Department of Signal Theory and Communications, Universidad Carlos III de Madrid, Madrid, Spain}

\icmlcorrespondingauthor{Aurora Cobo Aguilera}{acobo@tsc.uc3m.es}
\icmlcorrespondingauthor{Antonio Artés-Rodríguez}{antonio@tsc.uc3m.es}
\icmlcorrespondingauthor{Pablo Martínez Olmos}{olmos@tsc.uc3m.es}

\icmlaffiliation{eth}{Swiss Data Science Institute (ETHZ/EPFL), Zurich, Switzerland}

\icmlcorrespondingauthor{Fernando Pérez-Cruz}{fernando.perezcruz@sdsc.ethz.ch}

\icmlkeywords{Variance Reducer, Regularization, Deep Learning, Convolutional Neural Networks, Mini-batch Selection}

\vskip 0.3in
]



\printAffiliationsAndNotice{}  

\begin{abstract}
Deep learning requires regularization mechanisms to reduce overfitting and improve generalization. We address this problem by a new regularization method based on distributional robust optimization. The key idea is to modify the contribution from each sample for tightening the empirical risk bound. During the stochastic training, the selection of samples is done according to their accuracy in such a way that the worst performed samples are the ones that contribute the most in the optimization. We study different scenarios and show the ones where it can make the convergence faster or increase the accuracy.
\end{abstract}

\section{Introduction}
\label{sec:introduction}

Machine learning algorithms assumed that the samples are coming iid from $p(\mathbf{x},y)$ and hence they use the samples equally during training. For example, in deep learning all the samples enter with the same probability in each of the mini-batches \cite{bengio}. But not all samples are equally relevant when learning classifiers and regressors. Because some samples might be hard or easy to classify or they might be under-sampled or over-sampled in the training set without our knowledge. There are many ways in which non-uniform sampling can be used to improve convergence speed or quality by relying on non-uniform sample. The first example that comes to mind is AdaBoost \cite{schapire1998boosting}, which uses different weights for each training example to build a robust classifier.

More recently there has been proposals to use importance sampling for training classifiers and regressors to reduce the variance of their estimates. In a nutshell, the objective is to increase the number of times a hard-to-learn sample appears in the mini-batch so the learning algorithm can converge faster and then weight their error by the number of times it has been used. For example in \cite{borsos2018online}, the authors developed a non-uniform importance sampling technique to solve an online optimization problem with bandit feedback. In \cite{namkoong2017adaptive} the authors used the data structure to adapt the gradients of each observation. And \cite{salehi2017stochastic} try to sample the datapoints from a non-uniform distribution according to a multiarmed bandit framework. 

In the recent award-winning \cite{varianceBased}, the authors proposed bounds to reduce the variance of classifiers by relying on non-uniform sampling of the training set, but they did not compensate for the over-sampling  (or under-sampling) of the training set in their bound. The non-uniform sampling is a feature that should make the learned classifier more robust and reduce the variance of its prediction. The results in \cite{varianceBased} are mainly theoretical and they illustrate their algorithm in an example with very few training samples and large input dimension and using a logistic-regression classifier. 

In this paper, we embark on an implementation of this algorithm for training deep learning models to understand if this theoretical result shows significant improvement for standard deep learning classifiers. We first propose two different alternatives on how to incorporate the non-uniform sampling within the mini-batches used in deep learning, leading to different ways in which hard-to-classify examples are repeated in the mini-batches. 

We then compare these algorithms with a standard optimization of neural networks. We have relied on standard architectures and datasets not to biased our results with new neural networks or data. We found that there are some minor improvements in the convergence speed and reduction of error, but those improvements are not statistically significant. Also, there is not a consistent setting for the hyper-parameters for our algorithms that always improves the baseline. The proposed algorithms does not seems to hurt either and their computational complexity is negligible compared to the training of the neural network.

Also, we have noticed that if we do not use dropout \cite{srivastava2014dropout}, the improvement from using non-uniform sampling is significant. The improvements gains provided by dropout are equivalent to those of using our proposed implementation for \cite{varianceBased}. Even though, both methods are thought for reducing the variance of the learnt models, they achieve comparable results by the completely different means.

\cite{csiba2018importance} proposed a similar application, unifying importance sampling and minibatching algorithms so to assign some probability distributions to the samples of a set of minibatches and sample them. They propose a sampling scheme to improve the converge rates but, unlike us, they use probabilities to sample more relevant examples.

In a Bayesian setting, non-uniform sampling has been proposed in \cite{wang2016robust}. In it, the authors took a probabilistic approach in order to make inference by raising the likelihood of each data point to a weight. But in this paper the authors assume that the hard-to-learn samples are outliers that would contaminate the solution of our classifier and the algorithm actually under-samples them. The goal of sampling in this case is to reduce the outliers and not to make the classifier more robust to hard-to-classify examples that are still valid samples.

The rest of the paper is outlines as follows. We review the main results in \cite{varianceBased} in Section \ref{sec:motivation} and the proposed algorithms are detailed in Section \ref{sec:methodology}. We then present extensive empirical results in Section \ref{sec:experiments}. We conclude the paper in Section \ref{sec:conclusions}.

\section{Motivation}
\label{sec:motivation}
\subsection{Variance-based robust regularization}
\label{sub:variance}
\cite{varianceBased} proposed an alternative to empirical risk
minimization that provides a robust and computationally efficient
solution for small data sets. Particularly, it is based on tightening
the empirical risk bound by adding a variance term in the form
\begin{equation*}
	\frac{1}{n} \sum_{i=1}^{n} l (\theta, x_i) + C \sqrt{\frac{2\rho}{n}\text{Var}_{\hat{P}_n}( l\left( \theta,X\right) }
\end{equation*}
where $l$ is loss function, $C$ is a parameter that depends on $l$ and
the desired confidence guarantee, and $\text{Var}_{\hat{P}_n}$
the empirical variance.
\subsubsection{The empirical risk extension}
\label{subsub:extension}
Instead minimizing this regularized risk functional, generally
not-convex, the authors define a robust regularized risk
\begin{equation*}
	R_n\left( \theta, \mathcal{P}_n\right) = \sup_{P \in
          {\cal{P}}_n} \left\{ \mathbb{E}_{P}\left[ l\left(
              \theta,X\right) \right]: D_\phi(P||\hat{P}_n)\leq \frac{\rho}{n} \right\} 
\end{equation*}
where $D_\phi$ is the $\phi$-divergence with $\phi(t)=1/2
(t-1)^2$. The robust regularized risk is shown to bo equivalent to
\begin{equation}
\label{eq0}
	R_n\left( \theta, \mathcal{P}_n\right) =  \mathbb{E}_{\hat{P}_n}\left[ l\left( \theta,X\right) \right] + \sqrt{\frac{2\rho}{n}\text{Var}_{\hat{P}_n}\left( l\left( \theta,X\right)\right) }+ \varepsilon_n\left( \theta \right) 
\end{equation}

		\subsubsection{A more intuitive formulation}
		\label{subsub:minmax}
		As the authors describe in their work, we can consider the \eqref{eq0} as a \textbf{min-max problem}, that is, an optimization with two steps.
			
		\begin{itemize}
			\item First, \textbf{the minimization of the weighted risk}, $\min\limits_{\theta}\frac{1}{n} \sum_{i=1}^{n} p_i l_i (\theta, x_i) $, where $\theta$ are the parameters to be computed, $\mathbf{x}$ is a set of $n$ samples and $p_i$ is the weight associated to each sample, so the samples with higher contribution to the loss function are the more valuable in the model.
			\item Second, \textbf{the maximization of the robust objective}, $\max\limits_p \sum_{i=1}^{n} p_i l_i$.
		\end{itemize}
			
		As a constraint, they propose the \eqref{eq1}, where $\rho$ is a parameter to select the confidence level. In the case that $p_i$ is equal to $1/n$ for every sample, the model would correspond to the empirical risk minimization, that is, all the samples have the same weight and indeed, the same contribution.
			
		\begin{equation}
		\label{eq1}
		p \in \mathcal{P}_n = \left\lbrace  p \in \mathbb{R}_{+}^{n} : \frac{1}{2}\left\| np - \mathbf{1} \right\|_2^2 \leq \rho, \langle \textbf{1}, p \rangle = 1  \right\rbrace 
		\end{equation}
			
		They give a number of theoretical guarantees and empirical evidences in order to show the optimal performance of the estimator with faster rates of converge and the improvement of out-of-sample test performance in different classification problems.
		
	\subsection{Application on Deep Learning}
	\label{sub:application}
	
	
	Nowadays, deep learning is known as a powerful framework for supervised learning \cite{bengio}. It allows the implementation of neural networks with as many layers and units as it is desired, providing a more or less sophisticated function to fit a specific dataset. The description of such algorithms is followed by the specification of a cost function, an optimization procedure and a model, what makes the robust objective proposed a direct application in the step of the risk minimization of this kind of tools.
	
	Moreover, neural networks sometimes require long training times when the graph architecture is some how complex. These methods require using all the data before updating the predictor. As a consequence, a small improvement at each iteration in the optimization could make huge differences in the performance at the end.
	
	In spite of the high capacity for the adaptation to complex models that deep neural networks have, they involve an excessive computational complexity that makes impossible to apply directly the two-step algorithm from \cite{varianceBased} summarized in Subsection \ref{sub:variance}. Evaluating gradients two times would imply go over the entire dataset two times per epoch. We have modified the algorithm in \cite{varianceBased} so its computational complexity when training neural networks is negligible compared to uniform sampling.  
	
	We would like to express the contribution of the variance to the upper bound of the empirical risk as a way of selecting more frequently the samples with more variance in the mini-batches of the neural networks.  This is equivalent to, in the step of computing the gradients, use more times the worse performed samples. With that choice we would like to sacrifice the common classes at better performance on the rare ones.
					
\section{Model description}
\label{sec:methodology}
This section describes the methods to select the samples of the mini-batch in a deep learning problem based on the idea described before. We propose four different algorithms.

\begin{algorithm}[t]
\caption{Variance Reducer per Mini-batch (VR-M)}
\label{alg:vrm}
\begin{algorithmic}[1]
\REQUIRE Datasets $\mathcal{D}_{\text{train}}$ and $\mathcal{D}_{\text{test}}$.
\ENSURE Test accuracy $\bm{\eta}$.
\STATE Initialize parameters $\bm{\theta}$, number of epochs $E$ and repetition rate $\epsilon$;
\FOR{$e = 1 \dots E$} 
\STATE Divide dataset $\mathcal{D}_{\text{train}}$ in $M$ mini-batches;
\FOR{$m = 1 \dots M$} 
\STATE $\left\lbrace \mathbf{x}_m, \mathbf{y}_m\right\rbrace \leftarrow$ Obtain next mini-batch $m$;
\STATE $\bm{\ell}_m \leftarrow$ Evaluate cross-entropy in mini-batch $m$;
\STATE $\bm{\theta} \leftarrow$ Update parameters with stochastic gradient descent (SGD);
\STATE  $\left\lbrace \mathbf{x}_{m+1}, \mathbf{y}_{m+1}\right\rbrace \leftarrow$ Substitute $\epsilon \cdot M$ samples with the $\left\lbrace \mathbf{x}_{m}, \mathbf{y}_{m}\right\rbrace $ of highest $\bm{\ell}_{m}$;
\ENDFOR
\STATE $\eta_e  \leftarrow$ Compute test accuracy on $\mathcal{D}_{\text{test}}$;
\STATE Shuffle  $\mathcal{D}_{\text{train}}$;
\ENDFOR
\end{algorithmic}
\end{algorithm}

In the first algorithm, we train the neural network in such a way that, at each iteration, we repeat a percentage of the worst performed samples from the previous mini-batch. This percentage will be a hyper-parameter of the model and has a similar role as the parameter $\rho$ in \eqref{eq1}, since it lets more samples to have more contribution. We refer to this algorithms as Variance Reducer per Mini-batch (VR-M) and it is described in Algorithm \ref{alg:vrm}. 

In the second algorithm, we modify the original training set for each epoch so that we repeat a percentage of worst performed samples from the training of the whole previous epoch. This algorithm is detailed in Algorithm \ref{alg:vre} and we denoted by Variance Reducer per Epoch (VR-E). In this context, we must differ the connotation of iteration, what we mean as the step from a mini-batch to the next one, with respect to a step between two epochs, which includes many iterations.

\begin{algorithm}[t]
\caption{Variance Reducer per Epoch (VR-E)}
\label{alg:vre}
\begin{algorithmic}[1]
\REQUIRE Datasets $\mathcal{D}_{\text{train}}$ and $\mathcal{D}_{\text{test}}$.
\ENSURE Test accuracy $\bm{\eta}$.
\STATE Initialize parameters $\bm{\theta}$, number of epochs $E$ and repetition rate $\epsilon$;
\STATE Initialize $\mathcal{D}^{(1)}_{\text{train}} = \mathcal{D}_{\text{train}}$;
\FOR{$e = 1 \dots E$} 
\STATE Divide dataset $\mathcal{D}^{(e)}_{\text{train}}$ in $M$ mini-batches;
\FOR{$m = 1 \dots M$} 
\STATE $\left\lbrace \mathbf{x}_m, \mathbf{y}_m\right\rbrace \leftarrow$ Obtain next mini-batch $m$;
\STATE $\bm{\ell}^{(e)}_m \leftarrow$ Evaluate cross-entropy in mini-batch $m$;
\STATE $\bm{\theta} \leftarrow$ Update parameters with Stochastic Gradient Descent (SGD);
\ENDFOR
\STATE $\eta_e  \leftarrow$ Compute test accuracy on $\mathcal{D}_{\text{test}}$;
\STATE Shuffle  $\mathcal{D}_{\text{train}}$;
\STATE   $\mathcal{D}^{(e+1)}_{\text{train}} \leftarrow  \mathcal{D}_{\text{train}}$;
\STATE  $\mathcal{D}^{(e+1)}_{\text{train}}  \leftarrow$ Substitute  $\epsilon \cdot E$ samples with $\left\lbrace x_i, y_i\right\rbrace \in \mathcal{D}^{(e)}_{\text{train}}$ of highest $\bm{\ell}^{(e)}$;
\ENDFOR
\end{algorithmic}
\end{algorithm}

While with the \textit{VR-M} we can repeat a sample almost every iteration, with the \textit{VR-E} we restrict the number of times that a sample is repeated in the overall training because we have much fewer epochs than iterations.

We modify these two algorithms by not including all the samples but only a subset of them. We apply a sampling step with a 50\% of random data points belonging to the selection of the top-ranking worst performed ones. This approach helps the method not to insist always on the same samples (which could degrade the quality of the system) and makes the model more robust. That is, if there is a sample that is misleading the method, we could avoid its permanent contribution to the gradients with this solution. In order to make reference to both scenarios it is used Probabilistic Variance Reducer per Mini-batch (PVR-M) and Probabilistic Variance Reducer per Epoch (PVR-E) respectively for the first and the second models. 

Making more clear the differences between this last approach and the basic one, we are exposing an example. Therefore, if in the first algorithm it is repeated at each mini-batch the 40 samples with higher value in the lost function, with the probabilistic approach it would be repeated 20 random samples from these 40 ones.

\begin{figure}[t]
\vskip -0.1in
	\begin{center}
	\includegraphics[width=8cm]{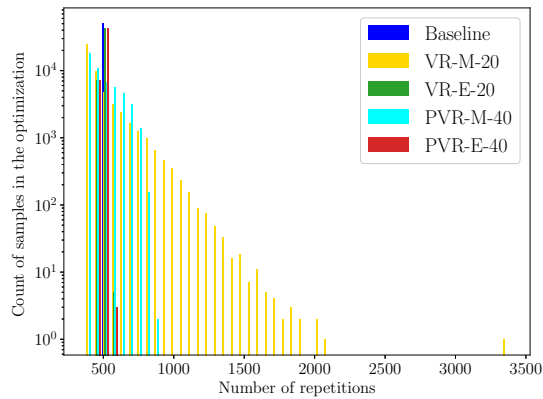}
	\caption{Histogram with the number of repetitions of the samples in the cifar-10 dataset with the all-cnn architecture. It is used a mini-batch of 128 samples and a dropout of 0.5. It is compared the percentages of 20 and 40 for both the model and the probabilistic approach.}
	\label{fig:histogram}
	\end{center}
\end{figure}

The 
 \ref{fig:histogram} shows an histogram with the number of times that a sample is used in the optimization. We compare the baseline, that is the original model without repeating any sample, with the two models and their probabilistic approaches. In order to have similar scenarios, we use a repetition of 20\% of the samples in the basic versions and 40\% with the probabilistic approaches. That is because the latter is resampled half its size so we retain just a quantity of 20\% of repeated samples at the end. Indeed, this idea is appreciated better in the Variance-Reducer per Epoch, with almost the same distribution of repeated samples, green and red color bars in the figure. Moreover, we can observe the idea mentioned before, that is, with the probabilistic approach we do not let the model to repeat a sample too many times, as it could happen with the model in yellow with a contribution of almost 3500 times from a set of samples. The baseline defines the number of iterations of the model, 500, that is the number of epochs, since a sample contributes one time per epoch in a original deep learning algorithm.

\section{Experiments}
\label{sec:experiments}
\subsection{Model}
\label{subsec:model}
We trained our method in a classification problem through several scenarios in order to generalize its properties. In consequence, we studied different datasets and networks from the literature.

\subsubsection{Datasets}
\label{subsubsec:data}
Between all the available datasets, it has been chosen the benchmark MNIST, SVHN and CIFAR-10 due to their multiple appearances in state-of-the-art works. They allow an easier and faster training of the experiments in comparison with larger bases as the ImageNet \cite{deng2009imagenet}. For this reason, the extension of this work in more complex domains will remain as a future task.

The MNIST is composed of 60000 training samples and 10000 test samples of handwritten digits \cite{lecun1998gradient}. The images are of size 28x28 pixels in gray scale. It is the simpler dataset used in this work.

The SVHN consists on 32-by-32 RGB images of house numbers from Google Street View \cite{netzer2011reading}. It has 73257 digits for training and 26032 digits for testing.

Finally, the CIFAR-10 collects labeled images of 10 classes \cite{krizhevsky2009learning}. They are 32x32 color pixels and a total of 50000 samples for training and 10000 for test.

\subsubsection{Architectures}
\label{subsubsec:arch}
As it was mentioned in the section \ref{sub:application}, we are proving the behavior of our method in CNNs. For this purpose, we are studying two different architectures of networks from the literature adapted to the datasets of section \ref{subsubsec:data}.

The first one is based on the VGG implemented by \cite{VGG}. The motivation of this choice is the validation of the method in a complex enough neural network where the improvements are considerably more cost efficient. The original architecture has been modified according to the size of our data, resulting in a neural network of 11 layers. It has three levels, the first one with two convolutional layers of output 16, the second with other two of output 32 and the third with four layers of output 64. All levels are ended with a max-pooling and finally it is applied three fully connected layers of size 1024, except the last one, with size the number of classes. This scheme is resumed in the table \ref{table:architectureVGG}.

\begin{table}[t]
\caption{Architecture VGG11b based on the VGG of 11 layers by \cite{VGG}.}
\label{table:architectureVGG}
\vskip 0.15in
\begin{center}
\begin{small}
\begin{sc}
\begin{tabular}{lc}
\toprule
Input image \\
\midrule
Conv 3x3-16 (with ReLu)  \\
Conv 3x3-16 (with ReLu)  \\
Max-pooling 2x2  \\
\midrule
Conv 3x3-32 (with ReLu)             \\
Conv 3x3-32 (with ReLu)      \\
Max-pooling 2x2       \\
\midrule
Conv 3x3-64 (with ReLu)               \\
Conv 3x3-64 (with ReLu)   \\
Conv 3x3-64 (with ReLu)   \\
Conv 3x3-64 (with ReLu)   \\
Max-pooling 2x2       \\
\midrule
Fully-Connected 1024 (with ReLu)               \\
Dropout 0.5       \\
Fully-Connected 1024 (with ReLu)               \\
Dropout 0.5       \\
Fully-Connected \#classes              \\
\midrule
Soft-max             \\
\bottomrule
\end{tabular}
\end{sc}
\end{small}
\end{center}
\vskip -0.1in
\end{table}

The second one is the network All-CNN-C from \cite{springenberg2014striving}. This particular architecture replaces the max-pooling choice by convolutional layers with increased stride as shown in the table \ref*{table:architectureALLCNN}. In the training of this scheme, it has been used an adaptive learning rate as in the original work.

\begin{table}[!th]
\caption{Architecture of the All-CNN-C by \cite{springenberg2014striving}.}
\label{table:architectureALLCNN}
\vskip 0.15in
\begin{center}
\begin{small}
\begin{sc}
\begin{tabular}{lcr}
\toprule
Input image \\
\midrule
Dropout 0.8       \\
Conv 3x3-96 (with ReLu) \\
Conv 3x3-96 (with ReLu) \\
Conv 3x3-96 (with ReLu) stride r=2 \\
\midrule
Dropout 0.5       \\
Conv 3x3-192 (with ReLu)             \\
Conv 3x3-192 (with ReLu)      \\
Conv 3x3-192 (with ReLu) stride r=2 \\
\midrule
Dropout 0.5       \\
Conv 3x3-192 (with ReLu)             \\
Conv 1x1-192 (with ReLu)             \\
Conv 1x1-\#classes (with ReLu)      \\
Global averaging over 6x6 spatial dimensions    \\
\midrule
Soft-max             \\
\bottomrule
\end{tabular}
\end{sc}
\end{small}
\end{center}
\vskip -0.1in
\end{table}

All the experiments have been trained with \textit{tensorflow}.

\subsection{Results}
\label{subsec:results}

The results shown in this section are trained through $200$ or $500$ epochs and with different distributions of the train and test sets, so we can notice one of the advantages of our work in the scenario with less training images. In the cases where we reduce the number of training samples, those ones that are removed are included in the validation set, so it will not be convenient to compare scores with different number of training images. In the tables \ref{table:results_mnist}, \ref{table:results_svhn} and \ref{table:results_cifar10} we resume the validation accuracy of different configurations and we remark in bold the scores that overcome the baseline and in red the best choice among all.

\begin{figure}[!th]
	\begin{center}
	\vskip -0.1in
		\subfigure[$60000$ training images.]{
			\includegraphics[width=8cm]{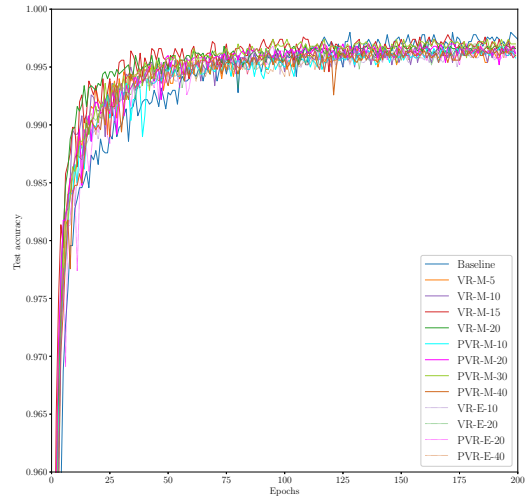}
			\label{fig:acc1}}
		\vskip -0.08in
		\subfigure[$50000$ training images.]{
			\includegraphics[width=8cm]{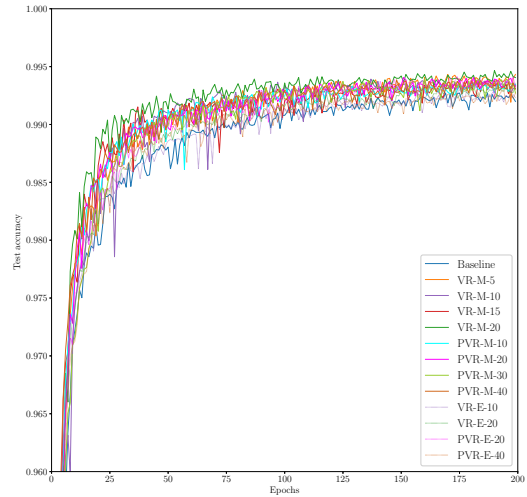}
			\label{fig:acc2}}
	\end{center}
	\caption{Validation accuracy per epoch in the MNIST dataset with the VGG11b architecture. It is used a mini-batch of 64 samples and a dropout of $0.5$. It is compared the percentages of samples repetition as detailed in the table \ref{table:results_mnist}.}
\end{figure}

In the case of the MNIST dataset, we used the VGG with 11 layers as described in table \ref{table:architectureVGG}. The mini-batch size was set to 64, the learning rate $0.001$ and the initialization of the parameters was $0.1$ for the standard deviation of the weights and $0.0$ for the biases. $200$ epochs were enough for all the scenarios to converge except for the one with $1000$ training samples that we used $500$ epochs. The table \ref{table:results_mnist} resumes the validation accuracy for different number of training samples, from the original configuration, $60000$ training images, until $1000$. In addition, we wanted to check the behavior of our method without dropout, what we have called `30000 DP1' in the table, since we used $30000$ training samples and set dropout to 1, that is the same as removing it in \textit{tensorflow}.

\begin{table*}[t]
\caption{Validation accuracy on MNIST with the VGG11 based network.}
\label{table:results_mnist}
\vskip 0.15in
\begin{center}
\begin{small}
\begin{sc}
\begin{tabular}{lcccccccc}
\toprule
\multirow{2}{*}{Models} & \multicolumn{6}{c}{\# training samples} \\
  & 60000 & 50000 & 40000 & 30000 & 20000 & 10000 & 1000 & 30000 DP1\\
\midrule
Baseline & \textcolor{red}{99.760\%} & 99.199\% & 99.046\% & 98.921\% & 98.626\% & 98.319\% & 93.934\% & 98.128\%\\
\midrule
VR-M-5          & 99.599\% & \textcolor{red}{\textbf{99.432\%}} & 99.018\% & \textbf{99.044\%} & \textbf{98.802\%} & 98.181\% & \textcolor{red}{\textbf{95.086\%}} & 97.899\%\\
VR-M-10        & 99.619\% & \textbf{99.312\%} & \textbf{99.099\%} & \textbf{99.064\%} & \textbf{98.722\%} & 98.259\% & 93.727\% & \textbf{98.134\%} \\
VR-M-15        & 99.659\% & \textbf{99.299\%} & \textcolor{red}{\textbf{99.207\%}} & \textbf{99.030\%} & \textbf{98.844\%} & \textbf{98.416\%} & \textbf{94.420\%} & \textbf{98.217\%}\\
VR-M-20       & 99.659\% & \textbf{99.406\%} & \textbf{99.123\%} & \textbf{99.061\%} & \textcolor{red}{\textbf{98.940\%}} & \textbf{98.414\%} & 93.411\% & \textbf{98.154\%}\\
\midrule
PVR-M-10      & 99.579\% & \textbf{99.346\%} & \textbf{99.203\%} & \textbf{98.953\%} & \textbf{98.809\%} & 98.254\% & \textbf{94.638\%}& \textcolor{red}{\textbf{98.355\%}}\\
PVR-M-20      & 99.619\% & \textbf{99.332\%} & \textbf{99.139\%} & \textbf{99.036\%} & \textbf{98.722\%} & 98.245\% & \textbf{94.258\%}& \textbf{98.177\%}\\
PVR-M-30      & 99.700\% & \textbf{99.272\%} & \textbf{99.163\%} & \textcolor{red}{\textbf{99.116\%}} & \textbf{98.829\%} & \textcolor{red}{\textbf{98.463\%}}& 93.616\% & 98.114\%\\
PVR-M-40      & 99.599\% & \textbf{99.306\%} & \textbf{99.187\%} & \textbf{98.998\%} & \textbf{98.800\%} & \textbf{98.443\%} & \textbf{94.272\%}& 98.060\%\\
\midrule
VR-E-10         & 99.599\% & \textbf{99.232\%} & \textbf{99.123\%} & 98.855\% & \textbf{98.691\%} & 98.248\% & \textbf{94.752\%} & \textbf{98.140\%}\\
VR-E-20         & 99.679\% & \textbf{99.359\%} & \textbf{99.111\%} & \textbf{98.978\%} & \textbf{98.637\%} & 98.142\% & \textbf{94.291\%} & \textbf{98.211\%}\\
\midrule
PVR-E-20      & 99.639\% & \textbf{99.319\%} & \textbf{99.099\%} & \textbf{98.998\%} & \textbf{98.729\%} & 98.250\% & 92.989\% & 97.897\%\\
PVR-E-40      & 99.679\% & \textbf{99.272\%} & \textbf{99.091\%} & 98.884\% & \textbf{98.717\%} & 98.172\% & 93.905\% & \textbf{98.292\%}\\
\bottomrule
\end{tabular}
\end{sc}
\end{small}
\end{center}
\vskip -0.1in
\end{table*}

We can conclude from the table \ref{table:results_mnist} that our model usually overcomes the baseline, even when we do not use another regularization mechanism as it is dropout. The unique configuration where we could not say any advantage a priori is the original configuration of the dataset, with $60000$ images, where the best score belongs to the baseline. However, we will see in the figure \ref{fig:acc1} that it is not like that and we can search for other interests in our method. Despite the fact that we have only presented one configuration without dropout because of the lack of time for completing more simulations, we can state through not shown tests that we can also obtain the same improvements in accuracy in other scenarios without dropout. Finally, regarding this dataset, the best model is the Variance-Reducer per Mini-Batch, although the advantages are obtained with both.

On the one hand, figure \ref{fig:acc1} exposes another interest of our method besides the improvement in the accuracy. That is the faster convergence. In the figure we can differentiate in blue the baseline curve that goes down the rest of the curves (several configurations of our method in the scenario with $60000$ training samples) in the first epochs, until the number $55$ approximately. After that, the convergence of the baseline follows a better score that the other ones. However, we could take use of this result to apply our method just in the first stage of the training in a particular problem so we can speed up the convergence.

On the other hand, figure \ref{fig:acc2} shows the accuracy evolution with $50000$ training samples, where our method works quite well, maintaining the baseline curve with the worst score during almost the complete training of the algorithm.

\begin{figure}[!th]
	\begin{center}
	\vskip -0.1in
	\includegraphics[width=8cm]{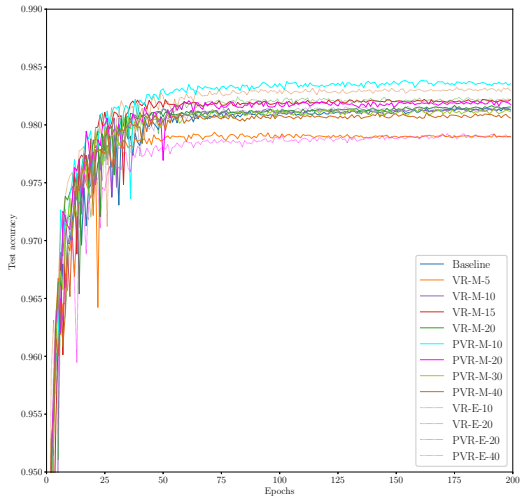}
	\vskip -0.3in
	\caption{Validation accuracy per epoch in the MNIST dataset with the VGG11b architecture. It is used a mini-batch of 64 samples and a dropout of $1$. It is compared the percentages of samples repetition as detailed in the table \ref{table:results_mnist}.}
	\label{fig:accDP}
	\end{center}

\end{figure}

Figure \ref{fig:accDP} shows the results for the experiments without dropout. In this case it is more visible the differences between the convergence of the baseline and our proposal methods. Precisely, the variance that each curve presents during the training is lower, what allows us to see them quite clear and distanced.

In the same way, we have trained the SVHN dataset with the same network and configuration than the MNIST, but with a batch size of $128$ and a fix number of $500$ epochs. The results are collected in the table \ref{table:results_svhn}.

\begin{table*}[t]
\caption{Validation accuracy on SVHN with the VGG11 based network.}
\label{table:results_svhn}
\vskip 0.15in
\begin{center}
\begin{small}
\begin{sc}
\begin{tabular}{lcccccccc}
\toprule
\multirow{2}{*}{Models} & \multicolumn{6}{c}{\# training samples} \\
 & 73257 & 60000 & 30000 & 20000 & 10000 & 5000 \\
\midrule
Baseline & 92.064\% & 91.753\% & 90.069\% & \textcolor{red}{88.787\%} & 80.938\% & \textcolor{red}{80.542\%}\\
\midrule
VR-M-5          & 91.649\% & 91.501\%  & 89.130\% & 86.933\% & \textbf{84.141\%} & 74.807\%\\
VR-M-10        & 90.602\% & 91.044\% & 88.258\% & 87.196\% & 80.813\% & 72.968\%\\
VR-M-15        & 91.364\% & 90.597\% & 88.254\% & 85.713\% & \textbf{81.606\%} & 65.123\%\\
VR-M-20       & 90.144\% & 90.403\% & 87.701\% & 86.561\% & \textbf{83.300\%} & 64.067\%\\
\midrule
PVR-M-10      & \textbf{92.372\% }& 91.557\%  & 89.474\% & 87.528\% & \textbf{83.941\%} & 72.991\%\\
PVR-M-20      & 91.918\% & \textbf{92.034\%}  & 89.542\% & 88.185\% & \textbf{85.252\%} & 66.201\%\\
PVR-M-30      & \textbf{92.438\%} & 91.621\%  & 89.497\% & 88.526\% &\textbf{ 83.444\%}& 73.885\%\\
PVR-M-40      & \textcolor{red}{\textbf{93.003\%}} & 91.577\%  & 89.688\% & 88.060\% & \textbf{85.246\%} & 73.303\%\\
\midrule
VR-E-10         & \textbf{92.153\%} & 91.590\%  & 89.760\% & 88.559\% & \textbf{84.522\%} & 78.712\%\\
VR-E-20         & \textbf{92.330\%} & \textbf{92.136\%}  & \textcolor{red}{\textbf{90.098\%}} & 87.901\% & \textbf{81.424\%} & 72.768\%\\
\midrule
PVR-E-20      & 91.687\% & \textcolor{red}{\textbf{92.175\%}}  & 89.613\% & 88.501\% & \textcolor{red}{\textbf{85.311\%}} & 74.320\%\\
PVR-E-40      & \textbf{92.403\%} & \textbf{91.804\%}  & 89.200\% & 88.559\% & \textbf{84.168\%} & 70.545\%\\
\bottomrule
\end{tabular}
\end{sc}
\end{small}
\end{center}
\vskip -0.1in
\end{table*}

The accuracy improvements for the classification of the images in the SVHN dataset are not so good. Not many models between the ones that we propose obtain better score than the baseline, except for the case with $10000$ training samples, where we overcome with until almost the $5\%$ the baseline score. When we use too few samples ($5000$ in the table \ref{table:results_svhn}) and the final accuracy is not high enough, the purpose of our method begins to lack of sense.

Finally, the training of the CIFAR-10 dataset is studied with the all-cnn  from table \ref{table:architectureALLCNN}. We employed a batch size of $128$ samples and the adaptive learning rate with the initial value of 0.01. The initialization of the parameters has been set to a standard deviation of $0.05$ for the weights and $0.0$ for the biases. In order to improve the baseline accuracy, we have applied a preprocessing step to the images that consists on a global contrast normalization and a ZCA whitening following \cite{goodfellow2013maxout}. The accuracies are exposed in the table \ref{table:results_cifar10}. 

With the study of this network, we can discover another possible advantage of our work in the configurations with less training samples. In the table \ref{table:results_cifar10}, when we decrease the number of training samples, our method works better and more cases that overcome the baseline appear. Moreover, the differences in the score between the baseline and the others are higher, so it has more sense the use of our approach with important improvement on the accuracy. That is something that happened in the SVHN dataset with $10000$ images, but this time in higher proportion with more than a $6\%$ of increase in the accuracy with $5000$ training images. Therefore, our method could be very useful when the number of samples is not high enough.

\begin{table*}[t]
\caption{Validation accuracy on CIFAR-10 with the all-cnn.}
\label{table:results_cifar10}
\vskip 0.15in
\begin{center}
\begin{small}
\begin{sc}
\begin{tabular}{lccccccc}
\toprule
\multirow{3}{*}{Models} & \multicolumn{6}{c}{\# training samples} \\
 & 50000 & 40000 & 30000 & 20000 & 10000 & 5000 \\
\midrule
Baseline & 88.131\% & \textbf{87.720\%} & 85.457\%& 82.545\% & 76.328\% & 69.360\% \\
\midrule
VR-M-5          & 87.981\% & 87.380\% & \textbf{85.763\%} & \textbf{83.246\%} & 76.232\% & 69.262\% \\
VR-M-10        & 88.041\% & 87.685\% & 85.266\% & \textcolor{red}{\textbf{83.777\% }}& \textbf{77.434\%} & \textbf{69.549\%} \\
VR-M-15        & 87.871\% & 87.565\% & 85.403\% & \textbf{82.562\%} & \textbf{76.899\%} & \textcolor{red}{\textbf{76.520\%}} \\
VR-M-20       & 88.061\% & 87.009\% & \textbf{85.677\%} & 82.537\% & \textbf{76.825\%} & 68.160\% \\
\midrule
PVR-M-10      & \textcolor{red}{\textbf{88.331\%}} & 87.309\% & \textcolor{red}{\textbf{85.991\%}}& \textbf{82.559\%} & \textbf{76.400\%} & \textbf{69.675\%} \\
PVR-M-20      & 88.131\% & 87.319\%  & 85.153\% & \textbf{83.188\%}& \textcolor{red}{\textbf{77.742\%}} & \textbf{70.161\%} \\
PVR-M-30      & 87.971\% & 87.354\% & 84.372\% & \textbf{83.213\%} & \textbf{77.220\%} & \textbf{69.537\%} \\
PVR-M-40      & 88.021\% & 87.650\% & 84.696\% & \textbf{82.762\%} & \textbf{77.003\%} & \textbf{69.639\%} \\
\midrule
VR-E-10         & 88.021\% & 87.019\% & 85.410\% & \textbf{82.933\%} & 76.117\% & 69.318\% \\
VR-E-20         & 87.720\% & 87.405\% & \textbf{85.577\%} & \textbf{82.802\%} & \textbf{76.446\%} & 69.169\% \\
\midrule
PVR-E-20      & 87.971\% & 86.899\% & \textbf{85.557\%} & \textbf{83.108\%} & 75.663\% & \textbf{69.668\%} \\
PVR-E-40      & 87.821\% & 86.859\% & 85.123\% & 82.379\% & \textbf{76.512\% }& \textbf{69.974\%} \\
\bottomrule
\end{tabular}
\end{sc}
\end{small}
\end{center}
\vskip -0.1in
\end{table*}

Regarding the choice of the percentage of repetition in the samples, we do not expose any evidence of trend that it may follow according to the number of training samples or the complexity of the network. Consequently, we should try different alternatives to find the best hyper-parameter. We just advise not to use very high percentages that would remove the sense of the method. Nevertheless, we found that the VR-M works better in dataset as MNIST and CIFAR-10, while in the case of SVHN, with the worse contribution of our method, VR-E appears to works better than the VR-M.

The code to launch the simulations from this section is released on GitHub\footnote{\href{https://github.com/AuroraCoboAguilera/RobustSampling}{github.com/AuroraCoboAguilera/RobustSampling}}.

\section{Conclusions}
\label{sec:conclusions}
In this work we have presented a novel idea for the selection of samples in the training of a deep learning model, based in the variance reduction of the real risk. It consists on the simple idea of repeating the samples with higher variance that are the ones with worse score in the cost function. 

We propose several models and study their performance in different architectures and datasets. We discuss the apparition of ones advantages and others according to the studied problem. Between them, we show the improvement of the accuracy in the classification, the faster rates of convergence and a better training when the number of samples is low. However, we do not expose any evidence for the choice of the value in the percentage hyper-parameter, what has to be tested in the problem to solve. Finally, we highlight the use of our work without dropout, with greater differences in the convergence accuracy and a more statistical relevant increase of the score.


\nocite{}

\bibliography{main}
\bibliographystyle{icml2019}


%


\end{document}